\pdfoutput=1
\documentclass[11pt]{article}

\usepackage{acl}

\usepackage{times}
\usepackage{latexsym}
\usepackage{amsfonts}

\usepackage[T1]{fontenc}
\usepackage[utf8]{inputenc}

\usepackage{graphicx}
\usepackage{subcaption}

\usepackage{microtype}
\usepackage{cinzel} %inserted by steven to add MERLOT font

\usepackage{multirow}
\usepackage{tabularx}

\usepackage{comment} %note: inserted by Steven for block commenting
\usepackage{wrapfig} %note: inserted by Steven for wrapping text around figures
\usepackage{calligra}

\usepackage{aurical}

\makeatletter
\def\thickhline{%
  \noalign{\ifnum0=`}\fi\hrule \@height \thickarrayrulewidth \futurelet
   \reserved@a\@xthickhline}
\def\@xthickhline{\ifx\reserved@a\thickhline
               \vskip\doublerulesep
               \vskip-\thickarrayrulewidth
             \fi
      \ifnum0=`{\fi}}
\makeatother

\makeatletter
\newcommand\footnoteref[1]{\protected@xdef\@thefnmark{\ref{#1}}\@footnotemark}
\makeatother

\newlength{\thickarrayrulewidth}
\newcommand*{\affmark}[1][*]{\textsuperscript{#1}}
\newcommand{\DatasetName}{\textit{PersonifCorp}}
\newcommand{\MethodName}{\textbf{\textcinzel{PINEAPPLE}}}

\newcommand{\titlecolor}{violet}
\title{\textcinzelblack{PINEAPPLE}: {\textcolor{\titlecolor}P}ersonifying {\textcolor{\titlecolor} I}{\textcolor{\titlecolor} N}animate {\textcolor{\titlecolor}E}ntities by {\textcolor{\titlecolor}A}cquiring {\textcolor{\titlecolor}P}arallel {\textcolor{\titlecolor}P}ersonification Data for {\textcolor{\titlecolor}L}earning {\textcolor{\titlecolor}E}nhanced Generation}

\author{Sedrick Scott Keh \affmark[1], Kevin Lu \affmark[2], Varun  Gangal\bf{\thanks{\quad Equal contribution by Varun and Steven}}{} \affmark[1], Steven Y. Feng\footnotemark[1]  \affmark[3], \\ 
\textbf{Harsh Jhamtani \affmark[1], Malihe Alikhani \affmark[4], Eduard Hovy \affmark[1]} \\ 
\affmark[1]Carnegie Mellon University, \affmark[2]University of Waterloo, \\ \affmark[3]Stanford University, \affmark[4]University of Pittsburgh \\  \texttt{\{skeh,vgangal,jharsh,hovy\}@cs.cmu.edu, syfeng@stanford.edu} \\
\texttt{kevin.lu1@uwaterloo.ca, malihe@pitt.edu}}

\begin{document}
\maketitle

\begin{abstract}
A personification is a figure of speech that endows inanimate entities with properties and actions typically seen as requiring animacy. 
In this paper, we explore the task of personification generation. To this end, we propose \textbf{\textcinzel{PINEAPPLE}}: \textbf{P}ersonifying \textbf{IN}animate \textbf{E}ntities by \textbf{A}cquiring \textbf{P}arallel \textbf{P}ersonification Data for \textbf{L}earning \textbf{E}nhanced Generation. We curate a corpus of personifications called {\DatasetName}, together with automatically generated de-personified literalizations of these personifications. We demonstrate the usefulness of this parallel corpus by training a seq2seq model to personify a given literal input. Both automatic and human evaluations show that fine-tuning with {\DatasetName} leads to significant gains in personification-related qualities such as animacy and interestingness. A detailed qualitative analysis also highlights key strengths and imperfections of {\MethodName} over baselines, demonstrating a strong ability to generate diverse and creative personifications that enhance the overall appeal of a sentence. \footnote{Data and code can be found at \url{https://github.com/sedrickkeh/PINEAPPLE}}
\end{abstract}

\section{Introduction}
\label{sec:intro}
Personification is the attribution of animate actions or characteristics to an entity that is inherently inanimate. Consider, for example, the sentence \textit{``The \textbf{stars danced playfully} in the moonlit sky.''} Here, the vibrance of the stars (something inanimate) is being likened to dancing playfully, which is a distinctly animate action. By allowing readers to construct clearer mental images, personifications enhance the creativity of a piece of text \cite{bloomfield1980personification,dorst2011personification,flannery2016personification}.

Being able to automatically identify and generate personifications is important for multiple reasons. First, humans naturally use personifications when communicating. When we say something like \textit{``My phone has died,''} or \textit{``My car is not cooperating,''} to a dialogue system, it is important that the dialogue system understands the intended meaning behind these personifications. If these systems interpret personifications literally, they may fail in several downstream tasks (e.g. classification) since their understanding is incorrect. Being able to generate personifications also allows dialogue agents and language models to be more creative and generate more figurative sentences. Personification generation has additional applications such as AI-assisted creative writing, since machine-generated figures of speech have been shown to enhance the interestingness of written text \cite{chakrabarty-etal-2021-mermaid}.

\begin{figure}
    \centering
    \includegraphics[width=\columnwidth]{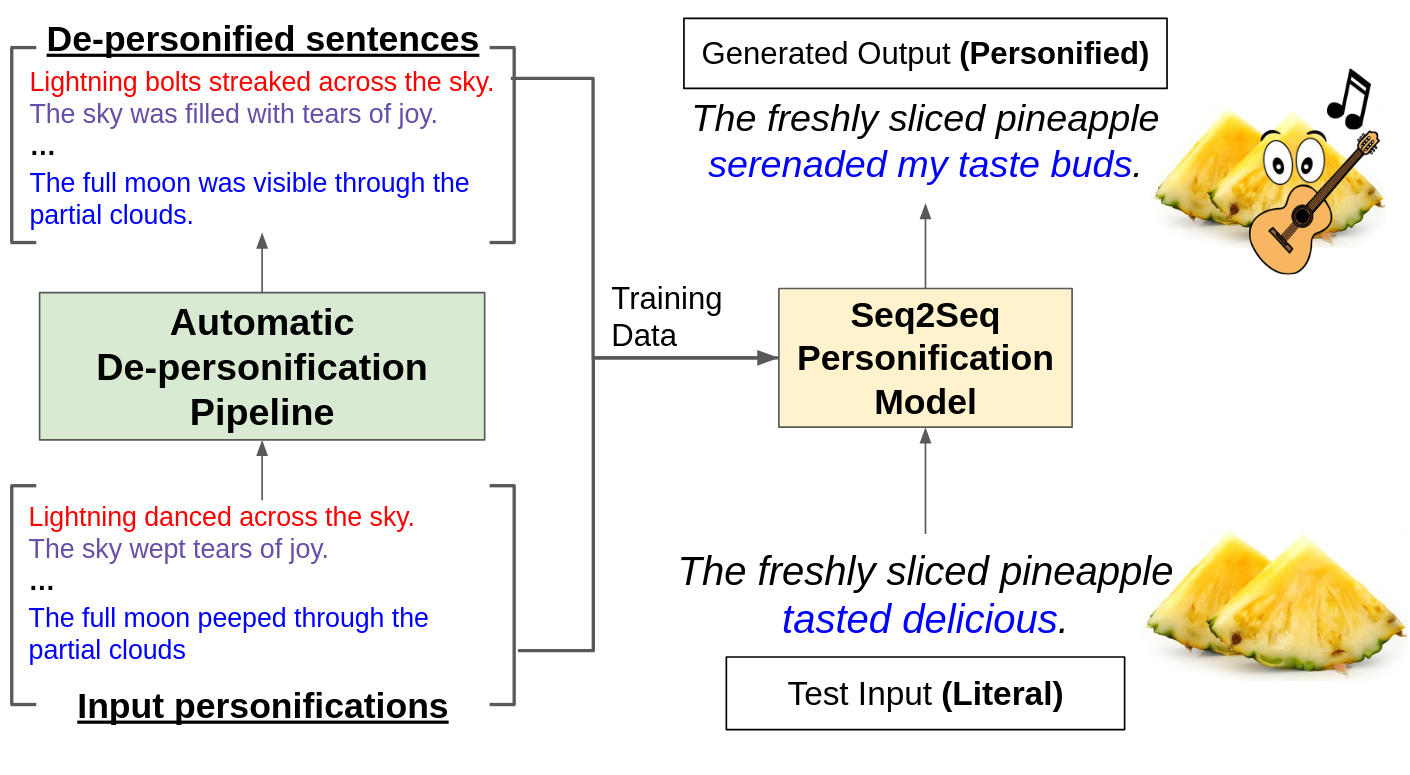}
    \caption{Overall {\MethodName} model pipeline. The left part of the diagram shows the corpus creation process, while the right part of the diagram shows the training and generation process.
    }
    \label{fig:main-diagram}
\end{figure}

Despite previous success in generating other figures of speech such as similes \cite{chakrabarty-etal-2020-generating}, metaphors \cite{stowe-etal-2021-metaphor}, hyperboles \cite{troiano-etal-2018-computational}, irony \cite{van-hee-etal-2018-semeval}, and sarcasm \cite{hazarika-etal-2018-cascade, jaiswal-2020-neural}, personification generation is relatively underexplored. One key challenge is that personifications do not have an explicit syntactic structure unlike similes which use \textit{`like'} or \textit{`as'}. They are also not as loosely-defined as metaphors. Rather, a personification requires identifying an inanimate subject together with actions or descriptions which are commonly used on animate subjects. These steps are challenging and require our models to understand commonsense concepts including animacy.

In line with exploring the task of personification generation, we present three main contributions:
(1) We curate a dataset, {\DatasetName}, of diverse personification examples from various sources. 
(2) We propose a method called {\MethodName} to automatically de-personify personifications and create a parallel corpus of personification data along with their literalizations. 
(3) Given our parallel corpus, we train a seq2seq model to personify given text. We conduct automatic and human evaluation and qualitative analysis of the generated outputs.

\section{Datasets}
\label{sec:dataset}
We curate a dataset called {\DatasetName} of 511 personifications, with 236 coming from a publicly available open-sourced list\footnote{\url{https://www.kaggle.com/datasets/varchitalalwani/figure-of-speech}} and 275 manually-filtered personifications extracted from the Deja dataset \cite{deja}. The Deja dataset is an image-captioning dataset containing a ``figurative'' subset of size 6000, of which $4.1\%$ of the captions are labelled as personifications. We extract these personifications and combine them with our existing list to form the final {\DatasetName} dataset.

We also note that although it is possible to further expand this dataset (e.g. by ad hoc searching for miscellaneous sites and examples online), we ultimately decide against this after performing an initial investigation. When we attempted to look for additional examples, we found that many of the new examples we found were near-duplicates of existing personifications already in our list. In addition, ad hoc searching can give at most a few hundred examples, which will lead to very incremental gains in performance. This is impractical if we want to collect a large-scale dataset. We hence decided to restrict ourselves to sentences from reasonably well-vetted, already existing corpora from *CL prior art or officially released data from sources like Kaggle/SemEval shared tasks.

\subsection{Characterizing Personifications}
We define the \textit{elements of personification}, an analogue to what was previously done for similes \cite{niculae-danescu-niculescu-mizil-2014-brighter, chakrabarty-etal-2020-generating}. While similes could be decomposed into very granular structures and well-defined elements, the unstructured nature of personifications prevents us from directly defining such fine-grained elements for personifications. Rather, we define two main high-level elements, the \textsc{topic} (a noun phrase that acts as logical subject) and the \textsc{attribute} (the distinctly animate action or characteristic that is being ascribed to the \textsc{topic}). Figure \ref{fig:pers-examples} shows examples of how these \textsc{topics} and \textsc{attributes} can relate to each other.

\subsection{Automatic Parallel Corpus Construction}
\label{sec:corpus-creation}
In order to train a seq2seq model to generate high-quality personifications, we need pairs of personifications along with their corresponding literalizations. However, the literalization process may take several human-hours, which is impractical for large datasets. We therefore propose {\MethodName}, a three-stage automatic de-personification process, where we first identify all valid \textsc{topic-attribute} pairs, then generate multiple candidates to replace the \textsc{attribute} of each \textsc{topic}. Lastly, we select the most appropriate candidate in terms of animacy, fluency, and meaning preservation. These steps are further detailed individually below:

\textbf{\textsc{Topic-Attribute} Extraction.} To identify the \textsc{topics} and \textsc{attributes}, we consider the dependency parse tree of a sentence and the part-of-speech (POS) tags of each of its words. Given the tree, we extract all the nouns/pronouns which have edges pointing into it with the \textit{nominal subject} label, together with the corresponding parent nodes. For instance, in the sentence \textit{``The stars danced in the night sky''}, the word \textit{`danced'} is a parent of the word \textit{`stars'}, with the \textit{nominal subject} edge relationship. We can thus identify \textit{`stars'} as the \textsc{topic} and \textit{`danced'} as the \textsc{attribute}. In more complex scenarios, we may need to perform some additional merging to deal with compound multi-word \textsc{topics} and \textsc{attributes}, as well as any additional modifiers. More specifically, using the POS tags, we identify all words tagged as \textit{negation modifiers}, \textit{possession modifiers}, \textit{nominal modifiers}, \textit{adjectival complements}, and \textit{objects of prepositions}, and words tagged as determiners and parts of compound phrases.\footnote{The spaCy library was used to extract the dependency tree and POS tags.} After extracting these nodes, they are iteratively merged with their parents in the dependency parse tree, and the merging process is performed repeatedly until no more merges are possible. The final \textsc{topic-attribute} pairs are then identified using the \textit{nominal subject} edge relationship as previously described. Examples of the merging process can be found in Appendix \ref{appendix:dependency-tree-merging}.

\begin{figure}[t]
\centering
\resizebox{1.0\columnwidth}{!} {
\begin{tabular}{cc}
  \begin{tabular}[t]{@{}c@{}} \textbf{\textsc{ATTRIBUTE}} \textbf{Type} \end{tabular} & 
  \textbf{Example} \\
 \hline \hline
 Noun & 
 \begin{tabular}[t]{@{}l@{}} The planet \textcolor{red}{earth}  is our \textcolor{blue}{mother}. \end{tabular} \\
 \hline
 Verb & 
 \begin{tabular}[t]{@{}l@{}} My \textcolor{red}{alarm clock} \textcolor{blue}{yells} at me to \\ get out of bed every morning. \end{tabular} \\
 \hline
 Adjective & 
 \begin{tabular}[t]{@{}l@{}} \textcolor{red}{Justice} is \textcolor{blue}{blind}  and, at times, \textcolor{blue}{deaf}. \end{tabular} \\
\hline 
\hline
\end{tabular}
}
\caption{Examples of different types of personification \textsc{attributes}  (\textsc{topics} in \textcolor{red}{red} and \textsc{attributes} in \textcolor{blue}{blue}).}
\label{fig:pers-examples}
\end{figure}

\begin{figure*}
    \centering
    \includegraphics[width=\textwidth]{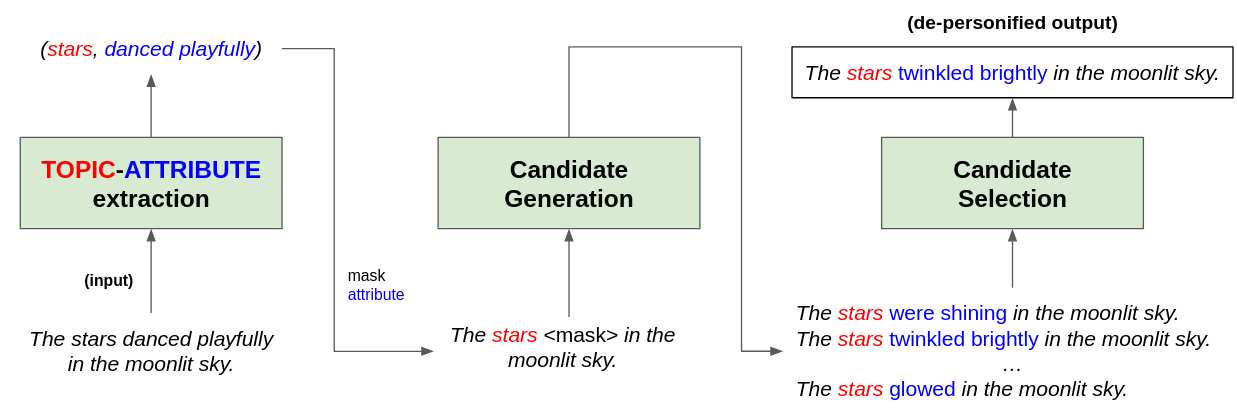}
    \caption{Overview of the {\MethodName} de-personification pipeline.}
    \label{fig:de-personification}
\end{figure*}

\begin{table*}[t]
    \centering \small
    \begin{tabular}{|p{0.45\textwidth}|p{0.45\textwidth}|}
    \hline 
         \textbf{Original Personification} & \textbf{Result After De-Personifying}  \\
         \hline \hline
         How far that little candle \textcolor{blue}{throws} its beams! & How far that little candle \textcolor{blue}{can spread} its beams! \\
         \hline
         A book is \textcolor{blue}{a fragile creature}, it \textcolor{blue}{suffers} the wear of time, it \textcolor{blue}{fears} rodents, the elements and clumsy hands. & A book is \textcolor{blue}{fragile}, it \textcolor{blue}{can break} from the wear of time, it \textcolor{blue}{can be eaten} by rodents, the elements and clumsy hands. \\
         \hline
         The camera \textcolor{blue}{loves} her since she is so pretty. & The camera \textcolor{blue}{is always on} her since she is so pretty. \\
         \hline
         Any trust I had for him \textcolor{blue}{walked} right out the door. & Any trust I had for him \textcolor{blue}{had gone} right out the door. \\
         \hline
         The full moon \textcolor{blue}{peeped} through partial clouds. &
         The full moon \textcolor{blue}{was visible} through partial clouds. \\
         \hline
         Opportunity \textcolor{blue}{was knocking} at her door. & 
         Opportunity \textcolor{blue}{was knocking} at her door. \\
         \hline
         The \textcolor{blue}{killing} moon will \textcolor{blue}{come} too soon. & 
         The killing moon will \textcolor{blue}{be here} too soon. \\
         \hline
    \end{tabular}
    \caption{Example outputs of the {\MethodName} de-personification pipeline. The \textsc{attributes} are highlighted in blue for both the original personifications, as well as the de-personified output sentences. The last two rows contain negative examples where the process does not successfully de-personify the input.}
    \label{tab:my_label}
\end{table*}

\textbf{Candidate Generation.}  
Once the \textsc{topic-attribute} pairs have been identified, we then determine which \textsc{topics} are inanimate. To achieve this, we need some type of commonsense notion of what constitues animacy. We use COMET \cite{bosselut-etal-2019-comet} to tap into the commonsense knowledge present in large-scale knowledge graphs such as ConceptNet \cite{Speer2017ConceptNet5A}. Although ConceptNet has no explicit notion of animacy, it has certain edge relations that we can leverage to design a proxy metric. More specifically, we use the \textit{IsA} relation to design a custom \textit{IsAPerson} animacy metric. If the \textsc{topic} of our sentence refers to an animate entity, then we expect its \textit{IsA} relation score with the word \textit{`human'} to be relatively low.\footnote{For the COMET ConceptNet graph, lower scores correspond to better matches.} The \textit{IsAPerson} metric is hence defined as follows: given a \textsc{topic}, we compute and average its \textit{IsA} scores to various words that are synonymous or very closely related to \textit{`human'}, such as \textit{`person'}, \textit{`man'}, and \textit{`woman'}. We call this set of \textit{`human'}-related words the \textsc{humanset}. The construction and full list of words in the \textsc{humanset} can be found in Appendix \ref{appendix:human-related-words}. The average of these ConceptNet scores is then our final  \textit{IsAPerson} animacy score.

Phrases whose \textit{IsAPerson} animacy score exceeds a certain threshold \footnote{We use a threshold of 7.0 for the \textit{IsAPerson} animacy metric. \textit{IsAPerson} scores $<$ 7.0 are considered animate, while scores $\geq$ 7.0 are considered inanimate. More details regarding the selection of this threshold can be found in Appendix \ref{appendix:params-and-thresholds}.}
are considered animate; otherwise, they are considered inanimate. Since our goal is to de-personify a sentence, we can safely discard all the animate \textsc{topics}, as these need no further de-personification. Rather, we focus on the inanimate \textsc{topics} because the segment we want to de-personify most likely occurs in the \textsc{topic-attribute} pairs whose \textsc{topic} is inanimate. Once we identify all such inanimate \textsc{topic-attribute} pairs, we mask out the \textsc{attribute} of each of them with \texttt{<mask>}, then use a pre-trained BART model \cite{lewis-etal-2020-bart} to generate the top $k=10$ candidates for each mask using beam search with a beam size of 10.
The goal of this process is to replace a possibly animate action/characteristic with candidates that are inanimate. 

\textbf{Candidate Selection.} Given $k=10$ candidate replacement \textsc{attributes}, we now select the most ideal replacement based on three metrics: animacy, fluency, and meaning preservation.
\begin{enumerate}
    \item Animacy -- We want the replacement \textsc{attribute} to be inanimate; otherwise we would just be replacing an animate \textsc{attribute} with another animate \textsc{attribute}. We define the animacy of a \textsc{topic-attribute} pair as difference between the affinity for a human ($\mathcal{A}_{human, \textsc{att}}$) to do/possess the \textsc{attribute}, and the affinity for the given \textsc{topic} ($\mathcal{A}_{\textsc{topic}, \textsc{att}}$) to do/possess the \textsc{attribute}. We use COMET's ConceptNet relations to compute these affinities; specifically, we use the \textit{CapableOf} relation. To approximate $\mathcal{A}_{human, \textsc{att}}$, we compute the average \textit{CapableOf} score between the given \textsc{attribute} and all words in our previously defined \textsc{humanset}. To compute $\mathcal{A}_{\textsc{topic}, \textsc{att}}$, we compute the \textit{CapableOf} score between the \textsc{topic} and its \textsc{attribute}. The final animacy score of a \textsc{topic-attribute} pair is defined as the difference $\mathcal{A}_{human, \textsc{att}} - \mathcal{A}_{\textsc{topic}, \textsc{att}}$.
    If there are multiple \textsc{topic-attribute} pairs, we consider the average animacy of all pairs.
    
    \item Fluency -- The de-personified sentences should be grammatically correct and sound natural. To measure for fluency, we use BART's generation scores (i.e. sum of individual token logits in the generated output).
    \item Meaning Preservation -- It is important that the de-personified sentence does not stray too far from the meaning of the original personification. We use BERTScore \cite{bert-score} between the de-personified and original sentences to measure meaning preservation.
\end{enumerate}
We design a composite scoring metric comprised of the aggregate scores from these 3 metrics.  
Due to scaling differences, we consider the $\log$ of the animacy score. To account for the fact that lower animacy scores imply less animate \textsc{topic-attribute} pairs (which is desirable in de-personification), we take the negative of the animacy. More precisely, we define our candidate score $S_i$ for candidate $i$ as 
$$S_{i} = \alpha \cdot (- \log(S_{anim.})) + \beta \cdot S_{flue.} + \gamma \cdot S_{mean.}$$
where $\alpha, \beta, \gamma$ are parameters. \footnote{We use $\alpha=1$, $\beta=1$, $\gamma=1$. Details about the tuning and selection of $\alpha, \beta, \gamma$ can be found in Appendix \ref{appendix:params-and-thresholds}.}  

Once $S_{i}$ is computed for all candidates, we select the candidate with the highest composite score as our final de-personified sentence. A diagram of the entire {\MethodName} pipeline is shown in Figure \ref{fig:de-personification}, and example outputs can be found in Table \ref{tab:my_label}.

\subsection{Test Data Construction}
While automatically generated pairs of personifications and literal de-personifications may greatly assist with training, these may not necessarily be accurate for testing. Rather, it would be more ideal during testing if we have ground-truth human-annotated data. To mimic our task at hand, we gather a list of non-personified English sentences.\footnote{\url{https://github.com/tuhinjubcse/SimileGeneration-EMNLP2020\#set-up-data-processing-for-simile}} We then select two annotators who are native English speakers currently enrolled in a university with English as a medium of instruction. 
These annotators were instructed to manually personify these sentences to create ground-truth reference personifications. The final {\DatasetName} test split has 72 literal + personified sentence pairs.

\section{Experimental Setup}
\label{sec:experimental_setup}
\subsection{Methods}
\label{sec:methods}
Below we outline the three models we consider, with two of them being naive baselines (COMET and Baseline-BART) that we simply use on {\DatasetName}'s test set, and the third (Finetuned-BART) being our proposed model trained on {\DatasetName}.

\begin{enumerate}
    \item \textbf{COMET}: We extract the \textsc{topic-attribute} pairs and identify the inanimate \textsc{topics} using the methods detailed in \S \ref{sec:corpus-creation}. Instead of generating candidate replacements using BART like in \S \ref{sec:corpus-creation}, we generate candidates by considering the top $k=10$ results for a given \textsc{topic} using COMET's ConceptNet \textit{IsCapable} relation (if the original \textsc{attribute} is a verb) or \textit{HasProperty} relation (if adjective or adverb). To incorporate a notion of animacy, we use the previously defined \textsc{attribute} animacy $\mathcal{A}_{human, \textsc{att}}$ and select the candidate with highest animacy as our final replacement.
    \item \textbf{Baseline-BART (BL-BART)}: We imitate the process outlined for the COMET baseline, except we use a pretrained BART model to generate the candidates instead of using COMET. All other steps (\textsc{topic-attribute} extraction and candidate selection) remain the same.
    \item \textbf{{\MethodName}-BART (PA-BART)}: We fine-tune a BART model by supplying the {\DatasetName} train split literal de-personified sentences (from the {\MethodName} pipeline) as inputs, and the original ground-truth personifications as target outputs. This is trained as a seq2seq task. During generation, we use beam search. Further details are outlined in \S \ref{sec:implementation-details}.  
\end{enumerate}

\subsection{Evaluation}
\label{sec:evaluation}
We consider both automatic evaluation metrics (\S \ref{sec:autoeval}) and human evaluation (\S \ref{sec:human-eval}).

\subsubsection{Automatic Evaluation}
For each model in \S \ref{sec:methods}, we evaluate its generated outputs on {\DatasetName}'s test split using each of the following automatic evaluation metrics:
\label{sec:autoeval}
\begin{enumerate}
    \item \textbf{BLEU} \cite{papineni2002bleu}: We use BLEU to ensure that the generations do not greatly differ from the inputs. We compute the BLEU score of each generated output with the literal inputs (for meaning preservation), as well as the ground-truth reference personifications.
    \item \textbf{BERTScore} \cite{zhang2019bertscore}: BERTScore measures how semantically related two sentences are, and is generally more robust
    than BLEU. We compute the BERTScore of each generated output with the inputs, as well as the ground-truth reference personifications.
    \item \textbf{Fluency}: To approximately measure the fluency of a sentence, we use generation (log-perplexity) losses of each output using the GPT-2 language model \cite{radford2019language}.
    \item \textbf{Animacy}: We are interested in how \textit{personified} the generated output is. We use the same animacy metric used for candidate selection in \S \ref{sec:dataset}, which is a combination of how animate the \textsc{attribute} is, as well as how inanimate the \textsc{topic} is. More precisely, this is defined as $\mathcal{A}_{human, \textsc{att}} - \mathcal{A}_{\textsc{topic}, \textsc{att}}$, where the $\mathcal{A}$ animacy scores are previously defined in \S \ref{sec:dataset}.
\end{enumerate}

\subsubsection{Human Evaluation}
\label{sec:human-eval}
The human evaluation was conducted using paid annotators on Amazon Mechanical Turk (AMT). Annotators were from Anglophone countries with $>97$\% approval rate.\footnote{More details about the human eval are in Appendix \ref{appendix:human-eval-setup}.}
Each test example was evaluated by exactly 2 annotators. For each test example, we first generate outputs using each of the methods outlined in \S \ref{sec:methods}. Corresponding to this test instance, we then create an AMT task page (a HIT), presenting the input literal sentence and each of the method outputs (in randomized order) for annotation along five aspects of text quality.

Specifically, annotation was elicited for the following metrics: \textbf{(1) Personificationhood} (\textit{``To what extent does the new sentence contain a personification?''}), \textbf{(2) Appropriateness} (\textit{``Do the personified nouns, verbs, adjectives, adverbs sound mutually coherent and natural?''}), \textbf{(3) Fluency} (\textit{``Does it sound like good English with good grammar?''}), \textbf{(4) Interestingness} (\textit{``How interesting and creative a rephrasing of the original sentence is the personified sentence?''}), and \textbf{(5) Meaning Preservation} (\textit{``Do the entities, their actions, interactions, and the events appear and relate to each other in the same way as in the original sentence?''}). Each metric was scored on a Likert scale, with 1 being the lowest and 5 being the highest.

For \textit{Interestingness}, we observed poor agreement scores amongst the AMT annotators.\footnote{Further details on inter-annotator agreement scores can be found in Appendix \ref{appendix:inter-annotator-agreement-scores}.}
Hence, for this aspect, we instead used a curated group of known, in-person annotators: a cohort of three native English-speaking students from an American university. Amongst these annotators, we observe a considerably higher agreement, with a Krippendorff $\alpha$ value of $0.5897$. For selecting this cohort from a slightly larger pool of candidates, we assessed their performance on a short qualification test of basic English literary skills and knowhow. The final cohort chosen each scored 85\% or higher on this test. Further details are in Appendix \ref{appendix:english-assessment}.

\subsection{Implementation Details}
\label{sec:implementation-details}
The {\DatasetName} training corpus was randomly split into a training and validation split with an 80-20 ratio.
We fine-tune a BART-base model with 139M parameters using
a learning rate of 2e-5 and a batch size of 4.
Training was done for 20 epochs and 400 warmup steps, and model/epoch selection was performed based on the lowest validation loss. For generating the outputs, decoding was done using beam search with a beam size of 10. Additional details can be found in Appendix \ref{appendix:implementation-details}.

\section{Results and Analysis}
\label{sec:results_and_analysis}
\subsection{Automatic Evaluation Results}

\begin{table*}[!t]
    \centering \small
    \begin{tabular}{c|cc|cc|c|c}
        \hline
        & \multicolumn{2}{c|}{\textbf{BLEU}} & \multicolumn{2}{c|}{\textbf{BERTScore}} & & \\
        & \textbf{Input} & \textbf{Gold} & \textbf{Input} &  \textbf{Gold} & \textbf{Fluency} $\downarrow$ & \textbf{Animacy} \\
        \hline \hline
        Human Annotation & 0.172 & 1.000 & 0.749 & 1.000 & 5.264 & 0.332 \\
        \hline
        COMET & 0.127 & 0.128 & 0.655 & 0.569 & 6.366 & -2.028 \\
        BL-BART & 0.132 & 0.133 & 0.728 & 0.617 & \textbf{4.573} & 0.106 \\
        PA-BART & \textbf{0.153} & \textbf{0.160} & \textbf{0.748} & \textbf{0.636} & 5.460 & \textbf{0.679} \\
        \hline
    \end{tabular}
    \caption{Average automatic evaluation results. The best-scoring method for each metric is highlighted in \textbf{bold}.
    Higher scores are better for all metrics except for fluency.} 
    \label{tab:auto-results-table}
\end{table*}

\begin{table*}[!t]
    \centering \small
    \begin{tabular}{c|c|c|c|c|c}
        \hline
        & \textbf{Personificationhood} & \textbf{Appropriateness} & \textbf{Fluency} & \textbf{Interestingness} & \textbf{Meaning Preservation}\\
        \hline \hline
        Human Annotation & 3.763 & 4.175 & 4.138 & 3.667 & 3.913 \\
        \hline
        COMET & 3.525 & 3.563 & 3.738 & 1.801 & 3.550 \\
        BL-BART & 3.500 & 3.938 & \textbf{4.188} & 2.006 & 3.750 \\
        PA-BART & \textbf{3.738} & \textbf{4.000} & 4.138 & \textbf{2.782} & \textbf{3.875} \\
        \hline
    \end{tabular}
    \caption{Average human evaluation results. The best-scoring method for each metric is highlighted in \textbf{bold}.
    }
    \label{tab:human-results-table}
\end{table*}

Table \ref{tab:auto-results-table} reports the automatic evaluation results for each of the metrics detailed in \S \ref{sec:autoeval}. We observe that our PA-BART model performs best across all automatic metrics except for fluency, 
where BL-BART performs best. The difference in performance is most significant in the \textit{Animacy} metric, which is the key metric that quantifies the degree to which a sentence is personified. This confirms that indeed, our proposed {\MethodName} method is successful in training a model to personify text. 

Our PA-BART model also performs well for both BLEU and BERTScore, scoring better than the COMET and BART baselines, and coming second only to the human-written personifications.

Lastly, with regards to fluency, the BL-BART model outperforms the PA-BART model. This is likely because when considering GPT-2 likelihood, it may unfavorably penalize creative sentences with personifications since these are naturally less common in regular text. As an example, the sentence \textit{``The stars danced playfully''} (GPT-2 loss = 7.02) would be deemed significantly less fluent than the sentence \textit{``The stars twinkled brightly''} (GPT-2 loss = 5.24), even though they are both valid sentences with similar meanings. This argument is further supported by the fact that even the reference human-generated personifications received a lower fluency score than the BL-BART outputs. Further, literal sentences are indeed typically more \textit{fluent} overall than personifications since they express the meaning literally. Nevertheless, we are still interested in the other qualities being measured by fluency: \textit{Is the sentence coherent? Does it make unnecessary grammatical errors?} In this regard, the fluency of PA-BART remains quite good. It is significantly better than the fluency of the COMET personifications 
and only slightly worse than the fluency of the human-written personifications. 

\subsection{Human Evaluation Results}
Human evaluation results are reported in Table \ref{tab:human-results-table}. 
Out of the five human evaluation metrics, the most pertinent metric to the personification generation task is \textit{Personificationhood}, as this metric explicitly tries to quantify the presence and overall quality of personifications. In this metric, our PA-BART model performs significantly better than both baselines and is only slightly worse than the human reference personifications. This indicates that PA-BART is very successful in generating personifications that humans are able to detect and understand.

Aside from measuring the presence of personifications, we also want to measure more fine-grained qualities of these personifications. This is done by considering the \textit{Appropriateness} and \textit{Interestingness} scores. In \textit{Interestingness}, PA-BART significantly outperforms both baselines but is worse than human annotations, while in \textit{Appropriateness}, PA-BART slightly outperforms BL-BART and is slightly worse than human annotations. Overall, we can conclude that the personifications generated by PA-BART are of good quallity: the \textsc{attributes} match up well with the \textsc{topics}, and they are overall very creative. This is further exemplified through the qualitative examples explored in \S \ref{sec:qualitative-analysis}.

Observations from \textit{Meaning Preservation} and \textit{Fluency} are very similar to those from the BLEU/BERTScore/Fluency metrics in the automatic evaluations. For \textit{Meaning Preservation}, PA-BART performs best among all models, and only slightly trails human references. Meanwhile, for fluency, BL-BART was ranked the most fluent, outperforming both PA-BART and the human references. As discussed previously, this is likely due to the fact that literal sentences are generally perceived to be more fluent than personifications.

\subsection{Qualitative Analysis}
\label{sec:qualitative-analysis}
\begin{table}[t]
\centering
\small
\addtolength{\tabcolsep}{-4pt}
\resizebox{\columnwidth}{!}{
\begin{tabular}{|p{46pt}|p{245pt}|}
\hline 
\textbf{Method} & \textbf{Text} \\ \hline
\hline
Literal Input & You are at a business dinner with your boss when your phone rings out loud (ex.1)\\ \hline
Human Ref & \textcolor{red}{You are at a business dinner with your boss when your phone starts singing out loud
} \\ \hline
COMET & \textcolor{orange}{You are at a business dinner with your boss when your phone beep out loud} \\ \hline
BL-BART & \textcolor{violet}{You are at a business dinner with your boss when your phone rings and you answer out loud} \\ \hline 
PA-BART & \textcolor{blue}{You are at a business dinner with your boss when your phone yells out loud} \\ \hline 
\hline
Literal Input & In most horror settings, silence is key. (ex.2)\\ \hline
Human Ref & \textcolor{red}{In most horror settings, silence is the protagonist.} \\ \hline
COMET & \textcolor{orange}{In most horror settings, silence scary.} \\ \hline
BL-BART & \textcolor{violet}{In most horror settings, silence is preferred.} \\ \hline 
PA-BART & \textcolor{blue}{In most horror settings, silence is a ghost.} \\ \hline 
\hline
Literal Input & Her relationships with family and friends are very difficult (ex.3)\\ \hline
Human Ref & \textcolor{red}{Her relationships with family and friends behave very unusually} \\ \hline
COMET & \textcolor{orange}{Her relationships with family and friends serious} \\ \hline
BL-BART & \textcolor{violet}{Her relationships with family and friends have always been strong.} \\ \hline 
PA-BART & \textcolor{blue}{Her relationships with family and friends are very lonely} \\ \hline 
\hline
Literal Input & Then there weren’t any more parties as the house became silent (ex.4)\\ \hline
Human Ref & \textcolor{red}{Then there weren’t any more parties as the house kept mum.} \\ \hline
COMET & \textcolor{orange}{Then there weren’t any more parties as the house build.} \\ \hline
BL-BART & \textcolor{violet}{Then there weren’t any more parties as the house fell into disrepair.
} \\ \hline 
PA-BART & \textcolor{blue}{Then there were no more parties as the house lamented.} \\ \hline 
\hline
Literal Input & It was a moonless nights, the air was still and the crickets were silent (ex.5)\\ \hline
Human Ref & \textcolor{red}{It was a moonless nights, the air was asleep and the crickets were silent} \\ \hline
COMET & \textcolor{orange}{It cold outside a moonless nights, the air cold outside still and the crickets noisy} \\ \hline
BL-BART & \textcolor{violet}{It was a moonless nights, the air was still and the crickets were calling.} \\ \hline 
PA-BART & \textcolor{blue}{It was one of those moonless nights, the air was tired and the crickets were silent
} \\ \hline 
\hline
Literal Input & The sound hit Frank loud enough to make your ear hurt (ex.6)\\ \hline
Human Ref & \textcolor{red}{The sound slapped Frank loud enough to make your ear hurt} \\ \hline
COMET & \textcolor{orange}{The sound echo Frank loud enough to make your ear sense sound} \\ \hline
BL-BART & \textcolor{violet}{The sound of Frank Sinatra is loud enough to make your ear ring.} \\ \hline 
PA-BART & \textcolor{blue}{The sound clapped loud enough to make your ear cry} \\ \hline 

\end{tabular}
}
\caption{\small Qualitative examples for personification: literal input, \textcolor{red}{human writing}, \textcolor{orange}{COMET}, \textcolor{violet}{BL-BART}, and \textcolor{blue}{PA-BART}. More can be found in Appendix \ref{appendix:more_qual_examples}.
}
\label{tab:qualitative_examples}
\end{table}
Table \ref{tab:qualitative_examples} contains a list of color-coded qualitative examples for each method. In Figure \ref{fig:pers-examples}, we previously outlined three main types of personification \textsc{topic-attribute} pairs, namely the cases where \textsc{attribute} is a noun, a verb, and an adjective. The first three examples in Table \ref{tab:qualitative_examples} demonstrate the capacity of our PA-BART model to capture all three cases. In the first example, the literal verb in ``\textit{your phone \textbf{rings} out loud}'' is replaced with the more appropriate animate verb in ``\textit{your phone \textbf{yells} out loud}.'' In the second, ``\textit{silence is \textbf{key}}'' is replaced with a noun in ``\textit{silence is \textbf{a ghost}}'', while in the third example, the literal adjective ``\textit{very \textbf{difficult}}'' is replaced with the animate adjective ``\textit{very \textbf{lonely}}''. These examples illustrate the generative flexibility of our model and its capacity to generate diverse outputs with different parts-of-speech.

We also observe that the outputs for PA-BART generally capture the meaning of the original text (and surrounding context) more accurately than the other baselines. In fact, the personifications greatly enhance the expressiveness of some of these sentences. In the first example, PA-BART replaces `\textit{rings}' with `\textit{yells}', while COMET replaces it with `\textit{beeps}', and BL-BART leaves `\textit{rings}' unchanged and just adds more details. Given the context of the sentence, we see that `\textit{yells}' is more appropriate, expressive, and consistent with the context. A similar argument can be made for most of the other examples in the table: 
for the third example, PA-BART replaces the literal \textit{``very \textbf{difficult}''} with the much more animate and expressive \textit{``very \textbf{lonely}''}, which is a suitable word to describe a relationship. In the fourth example, the BL-BART model is able to successfully capture the meaning of \textit{``the house \textbf{became silent}''} with \textit{``the house \textbf{fell into disrepair}''}. Although the meaning is correct, \textit{``fell into disrepair''} is more literal and does not contain a personification.
Compare this with the PA-BART's choice to replace \textit{``the house \textbf{became silent}''} with \textit{``the house \textbf{lamented}''}, which fits with the overall context (\textit{``Then there were no more parties...''}), and also greatly enhances creativity by invoking the vivid image of lamentation. Meanwhile, in the fifth example, BL-BART personifies \textit{``the crickets were \textbf{silent}''} with \textit{``the crickets were \textbf{calling}''}. However, this shift completely changes the meaning, so it is a rather poor choice of personification. In contrast, PA-BART rewrites \textit{``the air was \textbf{still}''} as \textit{``the air was \textbf{tired}''}, which is a reasonable personification that is consistent with the imagery in the sentence (\textit{``moonless nights''}, \textit{``crickets were silent''}). Hence, we see that PA-BART can generate creative and meaningful personifications, while simultaneously staying true to the spirit of the sentence.

We also point out that our model is not limited to single-word substitutions. Rather, it considers a holistic view of the entire sentence and modifies key segments as necessary. This allows PA-BART to handle compound phrases well: consider, for instance, the one-to-many-word substitution of \textit{`key'} $\longrightarrow$ \textit{`a ghost'} (example 2), and the many-to-one-word substitution of \textit{``became silent''} $\longrightarrow$ \textit{``lamented''} (example 4). More importantly, PA-BART is also able to simultaneously generate personifications in two disjoint parts of the sentence, as seen in the last example: ``\textit{The sound \textbf{clapped} loud enough to make your ear \textbf{cry}}.'' Here, there are two personifications in \textit{``sound \textbf{hit}''} $\longrightarrow$ \textit{``sound \textbf{clapped}''}, and \textit{``ear \textbf{hurt}''} $\longrightarrow$ \textit{``ear \textbf{cry}''}.

This last example also demonstrates the imperfection of our method. Although the model is able to generate two personifications, it loses a component of the original sentence because the recipient of the action (\textit{`Frank'}) has disappeared. This same issue of meaning or information loss is present in example 2, where our model's output of \textit{``silence is \textbf{a ghost}''}, while a personification, actually contradicts the original text \textit{``silence is \textbf{key}''}. BL-BART's output of \textit{``silence is \textbf{preferred}''}, while not a personification, correctly preserves the original meaning, as does the human reference of \textit{``silence is \textbf{the protagonist}''}. This suggests that the model may still need some improvements with balancing creativity and semantic preservation. Other possible weaknesses are outlined in \S \ref{sec:conclusion_future_work}.

\subsubsection{Novelty and Diversity Analysis}
We randomly sample 30 examples from the PA-BART generations- and manually identify the parts of the sentences that were personified, as well as the animate \textsc{attributes} used to personify the \textsc{topics}. Among the 30 examples, there were 27 unique \textsc{attributes}, and only 3 repeats. Additionally, there were 9 examples which generated completely new \textsc{attributes} that were never before seen in the training set, which demonstrates that the model is able to sufficiently capture the essence of a personification, rather than just blindly memorizing \textsc{attributes} from the training data.

\section{Related Work}
\label{sec:related_work}
We present the linguistic underpinnings behind the \textsc{topic-attribute} framework used in this paper and explore how other types of figures of speech are generated. We also explore 
what makes personification generation so challenging. 

\textbf{Linguistic Motivations.} Personifications traditionally do not have clearly defined classifications. In fact, even within the linguistic community, the definition of a personification is not always very clear-cut \cite{edgecombe-personification, hamilton-personification}. A study by \citet{Long2018MeaningCO} examines the personification structure ``\textit{nonhuman subject + predicate verb (used for humans only) + others},'' as well as the structure ``\textit{others + predicate verb (used for humans only) + nonhuman object + others}.'' We generalize and repackage these concepts, renaming the \textit{subject} as the \textsc{topic} and the \textit{predicate verb} as the \textsc{attribute}. In doing so, we are able to capture more general notions of animacy beyond just verbs.

\textbf{Generation of Metaphors, Similes, etc.} A lot of studies on metaphors have focused on identification using techniques like word sense disambiguation \cite{birke-sarkar-2007-active}, topic modeling \cite{strzalkowski-etal-2013-robust, heintz-etal-2013-automatic}, dependency structures \cite{jang-etal-2015-metaphor}, and semantic analysis \cite{hovy-etal-2013-identifying}. In terms of generation, early systems have explored grammar rules \cite{genmeta}, while more recently, large language models have greatly aided in metaphor generation. Most notably, \citet{stowe-etal-2021-metaphor} generate metaphors by considering conceptual mappings between certain domains and verbs. \citet{chakrabarty-etal-2021-mermaid} further build on this by creating a parallel corpus of metaphors and training a large language model to perform the generation.

We also note here that the two aforementioned studies already cover personifications to a certain extent. However, these studies considered personifications as subtypes of metaphors. Some of the methods used may not generalize well to other types of personifications. Our study is the first to focus specifically on generating personifications.

For generating similes, \citet{chakrabarty-etal-2020-generating} propose using style-transfer models with COMET commonsense knowledge to generate similes. The study similarly creates a parallel corpus and trains a seq2seq model to perform the generation.

There is also a recent work by \citet{sedrick_tongue_twisters} that uniquely investigates the generation of tongue twisters using seq2seq and language models.

\textbf{Personifications.} There are currently few studies that specifically work on personifications. \citet{gao-etal-2018-neural} detect personifications as a subtype of metaphors, but not as its own figure of speech. Generation is largely unexplored. We believe this is likely because personifications are generally more difficult to define and categorize. Furthermore, because several sources simplify personifications to fall under metaphors \cite{stowe-etal-2021-metaphor, chakrabarty-etal-2021-mermaid}, there is also a lack of personification-specific datasets.

\textbf{Constrained Text Generation.} There is also a body of work exploring the family of more general constrained text generation tasks. \citet{gangal2021nareor} investigate NAREOR, or narrative ordering, which rewrites stories in distinct narrative orders while preserving the underlying plot. \citet{miao2019cgmh} show gains on several tasks through determining Levenshtein edits per generation step using Metropolis-Hastings sampling.
\citet{feng2019keep} propose Semantic Text Exchange to modify the topic-level semantics of a piece of text.

\citet{lin-etal-2020-commongen} propose CommonGen, a generative commonsense reasoning task based on concept-to-text generation. Works investigating this task include EKI-BART \cite{fan2020enhanced} and KG-BART \cite{liu2020kg}, which use external knowledge to enhance performance on CommonGen. SAPPHIRE \cite{feng-etal-2021-sapphire} uses the data itself and the model's own generations to improve CommonGen performance, while VisCTG \cite{feng_caption} uses per-example visual grounding.

\section{Conclusion and Future Work}
\label{sec:conclusion_future_work}
In this paper, we explored the task of personification generation. We curated a dataset of personifications and proposed the {\MethodName} method to automatically de-personify text. Using our parallel corpus, \textit{PersonifCorp}, we trained a seq2seq model (BART) to generate creative personifications. Through automatic, human, and qualitative evaluation, we demonstrated that these personifications make sentences more interesting and enhance the text's overall appeal.
Our finetuned model successfully does this while maintaining a high level of fluency and meaning perservation.

Some weaknesses of our model include failing to personify more complex sentence structures, and occasionally failing to preserve the exact meaning of the original sentence. We also believe that our model still has room to grow in terms of the diversity of personifications generated. Further, we can explore unsupervised style transfer methods \cite{10.5555/3327757.3327831, malmi-etal-2020-unsupervised, style20}, where we regard the personificationhood of a sentence as a kind of style. We can also investigate data augmentation methods \cite{feng-etal-2021-survey, feng-etal-2020-genaug, dhole2021nl} to further expand our dataset. Another promising direction would be to explore ways to acquire more control over which parts of the sentence are personified or what types of personifications are generated, or to apply this to make dialogue agents more interesting, e.g. by giving them more personality \cite{Li_Jiang_Feng_Sprague_Zhou_Hoey_2020}.

\clearpage
\section*{Ethics Statement}
Our human and automatic evaluations (see \S\ref{sec:evaluation}) are done over content either directly sourced from, or generated by publically available, off-the-shelf pretrained models trained either on already existing, publicly available datasets, or datasets further derived by post-processing the same --- as further described in Datasets (see \S \ref{sec:dataset} for more).

We do collect human evaluation ratings using crowd-sourcing, specifically through AMT and in-person annotation. However, we neither solicit, record, nor request any kind of personal or identity information from the annotators. Our AMT annotation was conducted in a manner consistent with terms of use of any sources and intellectual property and privacy rights of AMT crowd workers. Crowdworkers were fairly compensated: \$1.12 per  fluency + appropriateness + meaning preservation evaluation HIT, and \$0.56 per personificationhood evaluation HIT, for roughly 6 min (first) and 2 min (latter) tasks, respectively. This is at least 1.5-2 times the minimum U.S.A. wage  of \$7.25 per hour (\$0.725 per 6 minutes and \$0.25 per 2 minutes).

We primarily perform experiments on personification in English \cite{bender2018data}.

NLG models are known to suffer from biases learnable from training or finetuning on data, such as gender bias \cite{dinan2020queens}. However, our work and contribution does not present or release any completely new model architechtures, and is primarily concerned with more careful adaptation and finetuning of existing pretrained models for a particular class of figurative construct (i.e. personification). The frailties, vulnerabilities, and potential dangers of these models have been well researched and documented, and a specific re-investigation would be repetitive and beyond the scope and space constraints of this paper.

We do not foresee any explicit way that malicious actors could specifically misuse fintuned models that could be trained on our data, beyond the well-researched, aforementioned misuse that is possible in general with their instantiation for any transduction task or dataset (e.g. summarization).

%\section*{Acknowledgements}

% Entries for the entire Anthology, followed by custom entries
\bibliography{anthology,custom}
\bibliographystyle{acl_natbib}

\newpage
\appendix
\section{Appendix A: De-Personification Pipeline}
\subsection{Dependency Tree Merging Example}
\label{appendix:dependency-tree-merging}
\begin{figure*}[t]
    \centering
    \includegraphics[width=\textwidth]{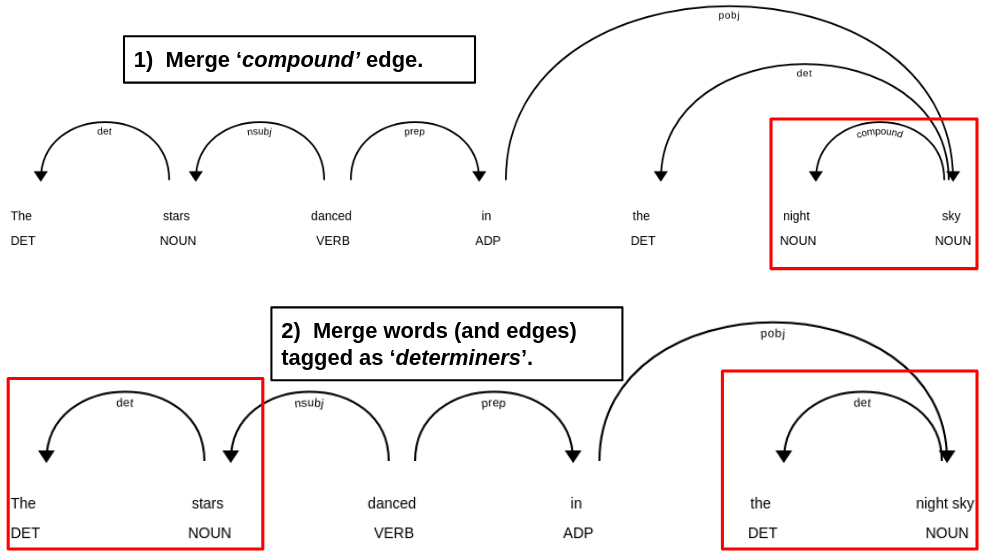}
    \includegraphics[width=0.9\textwidth]{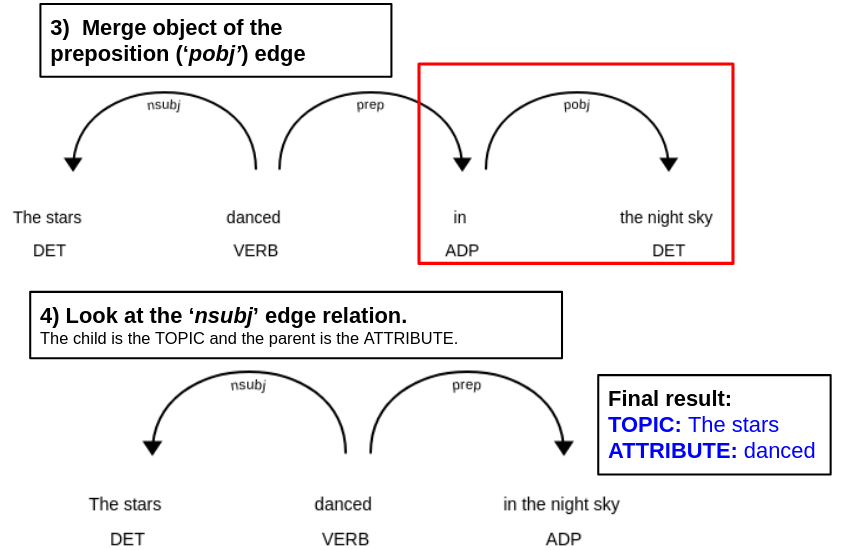}
    \caption{Step-by-step example of the merging process for \textsc{topic-attribute} identification.}
    \label{fig:merging}
\end{figure*}

Figure \ref{fig:merging} contains an example of the merging process that was described in the \textsc{topic-attribute} extraction step in \S \ref{sec:dataset}. As outlined in \S \ref{sec:dataset}, edge relations to iteratively merge are \textit{negation modifiers}, \textit{possession modifiers}, \textit{nominal modifiers}, \textit{adjectival complements}, and \textit{objects of prepositions}, as well as words tagged as determiners and parts of compound phrases. The priority order for merging is as follows: 1) compound phrases, 2) nominal modifiers, 3) possession modifiers, 4) negation modifiers, 5) determiners, 6) objects of prepositions, 7) adjectival complements.

\subsection{Human-Related Words}
\label{appendix:human-related-words}
In \S \ref{sec:dataset}, we defined the \textit{IsAPerson} animacy metric as the average of the \textit{IsA} scores between the \textsc{topic} and various words that are very closely related to \textit{`human'}. We called this set the \textsc{humanset}. The words contained in \textsc{humanset} are as follows: \{``person'', ``human'', ``man'', ``woman'', ``human being'', ``boy'', ``girl''\}.

These words were empirically selected by considering a list of synonyms of the word \textit{`person'} and checking the \textit{IsA} relation COMET scores with the word \textit{`human'}. All of the above words have \textit{IsA} scores with \textit{`person'} of less than 5.10.

\subsection{Parameters and Thresholds}
\label{appendix:params-and-thresholds}
\textbf{\textit{IsAPerson} Threshold. } For the \textit{IsAPerson} animacy score, we use a threshold of 7.0. \textit{IsAPerson} scores $< 7.0$ are considered animate, while scores $\geq 7.0$ are considered inanimate. This threshold was selected empirically using words known to be animate and words known to be inanimate. Words tested include ``she'' (5.31), ``person'' (6.41), ``moon'' (8.743), ``opportunity'' (9.488), ``stars'' (8.717), ``joe'' (5.804), ``jane'' (4.976), ``the police officer'' (6.462), ``my friend'' (6.805), ``my new iphone'' (10.055). From these observations, we observe that all animate words have an \textit{IsAPerson} score of $< 7.0$, while all inanimate objects have a score of $\geq 7.0$. We hence conclude that $7.0$ is a suitable threshold.

\textbf{Candidate Selection Composite Score Parameters. } For the $\alpha, \beta, \gamma$ used in the composite score for candidate selection, we use values of $\alpha=1, \beta=1, \gamma=1$. This was selected for two reasons. First, all of the score values had largely similar scales (logarithmic), so setting $\alpha, \beta, \gamma$ to a larger value like 2 or 3 would disproportionately favor a certain metric, which is not what we desire. Second, we experimented with using values such as 0.8, 1.2, and 1.5, but the generated de-personifications were either very similar or slightly worse than the default setting of $\alpha=1, \beta=1, \gamma=1$. A possible future direction would be to explore possible values of $\alpha, \beta, \gamma$ more thoroughly, but for this dataset, we stick to the simple case of $\alpha=1, \beta=1, \gamma=1$.

\section{Appendix C: Evaluation Details}
\subsection{Human Evaluation Setup}
\label{appendix:human-eval-setup}
A total of 20 unique AMT annotators participated in the study for fluency, appropriateness, and meaning preservation, each performing 4.0 HITs on average. Annotators were compensated 1.12\$ per HIT, each of which was designed to take <6 mins on average.

22 unique AMT annotators participated in the second, separate study for personificationhood, each performing 4.36 HITs on average. Annotators were compensated 0.56\$ per HIT, each of which was designed to take <2 mins on average.

For the interestingness study, the details regarding annotator background and selection can be found in \S \ref{sec:human-eval} and Appendix \ref{appendix:english-assessment}.

The html templates including instructions, questions and other study details corresponding to both these AMT studies can be found in the \texttt{templates/} subfolder of our code submission zip, with the names \texttt{fluency\_appropriateness\_} \texttt{meaningPreservation.html} and \texttt{personificationhood.html} respectively. 

\subsection{Inter-Annotator Agreement Scores}
\label{appendix:inter-annotator-agreement-scores}
\begin{table}[t]
    \small
    \centering
    \begin{tabular}{c|c|c}
         \textbf{Metric} &  \begin{tabular}[t]{@{}c@{}} \textbf{Spearman} \\ \textbf{Correlation} \end{tabular} & \textbf{Krippendorff $\alpha$} \\
         \hline
         Personificationhood & 0.0934 & 0.0250 \\
         Appropriateness & 0.1660 & 0.1778 \\
         Fluency & 0.0050 & 0.0942 \\
         Interestingness & 0.6160 & 0.5898 \\
         Meaning Preservation & 0.0389 & 0.2558 \\
    \end{tabular}
    \caption{Inter-annotator agreement scores.}
    \label{tab:inter-annotator-scores}
\end{table}

Each generated input instance and its respective model outputs are labelled by two distinct annotators. To measure inter-annotator agreement, we use Spearman correlation and Krippendorff $\alpha$, as reported in Table \ref{tab:inter-annotator-scores}.

To get the Spearman correlation point value for a given aspect and test instance, we compute mean pairwise Spearman correlation between the aspect values assigned to the corresponding model outputs by every pair of annotators. Specifically, we use the \textit{scipy.stats} implementation to compute this. \footnote{\url{https://docs.scipy.org/doc/scipy/reference/generated/scipy.stats.spearmanr.html}}

For Krippendorff $\alpha$, we treat each human evaluation aspect as an ordinal quantity. Specifically, we use the implementation provided by the python library \textit{krippendorff 0.5.1}.\footnote{\url{https://pypi.org/project/krippendorff/}}

\subsection{English Assessment Test for Annotators}
\label{appendix:english-assessment}
From the native English-speaking university student annotators who enrolled to participate in our Interestingness study, we first elicited answers to an English assessment test, as mentioned in \S\ref{sec:human-eval}.

The assessment test comprised of 12 questions spanning multiple  question types testing the examinee's grasp of the use and distinction between various figures of speech, basic literary general knowledge, and verbal reasoning skills. A spreadsheet file containing this test can be found with the name \textit{LiteratureTest.xlsx} under the \textit{Templates/} subfolder of our code submission .zip file.

The final annotators used for our interestingness study were chosen from those who got 11 or more of the 12 questions on the English assessment test correct, hence scoring at least $85\%$ on the test.

\section{Appendix B: Implementation Details}
\label{appendix:implementation-details}
The BART-base model was trained using a learning rate of 2e-5. This was by conducting a hyperparameter search over the values \{1e-6, 5e-6, 1e-5, 2e-5, 5e-5, 1e-4\} and selecting the model/epoch based on lowest validation loss. The same process was done to select a batch size of 4 using a hyperparameter search over values \{2,4,8,16\}. Training was done for 20 epochs and 400 warmup steps. The Adam optimizer was used, and inputs were truncated to a maximum length of 64 tokens (using BART's subword tokenization).

Training was done on Google Colaboratory environments using V100 GPUs. For the BART-base model, a single training loop of 20 epochs takes approximately 10 minutes to complete.

\section{Appendix D: Additional Examples}
\label{appendix:more_qual_examples}
Table \ref{appendix:qualitative_examples} is an extension of Table \ref{tab:qualitative_examples} and contains additional qualitative examples of the generations.

\begin{table}[t]
\centering
\small
\addtolength{\tabcolsep}{-4pt}
\resizebox{\columnwidth}{!}{
\begin{tabular}{|p{46pt}|p{245pt}|}
\hline 
\textbf{Method} & \textbf{Text} \\ \hline
\hline
Literal Input & The news hit me hard. (ex.7)\\ \hline
Human Ref & \textcolor{red}{The news punched me hard.} \\ \hline
COMET & \textcolor{orange}{The news report event late me hard.} \\ \hline
BL-BART & \textcolor{violet}{The news hit me hard.} \\ \hline 
PA-BART & \textcolor{blue}{The news pounded me hard.} \\ \hline 
\hline
Literal Input & Panic is sweeping through the streets contagiously. (ex.8)\\ \hline
Human Ref & \textcolor{red}{Panic is driving around the streets contagiously.} \\ \hline
COMET & \textcolor{orange}{Panic is result through the streets contagiously.} \\ \hline
BL-BART & \textcolor{violet}{Panic is running through the streets contagiously.} \\ \hline 
PA-BART & \textcolor{blue}{Panic runs through the streets contagiously.} \\ \hline 
\hline
Literal Input & In the dead of summer, in Ohio, the clothes felt hot and humid. (ex.9)\\ \hline
Human Ref & \textcolor{red}{In the dead of summer, in Ohio, the clothes seemed to be sweating themselves.} \\ \hline
COMET & \textcolor{orange}{In the dead of summer, in Ohio, the clothes fit and humid.} \\ \hline
BL-BART & \textcolor{violet}{In the dead of summer, in Ohio, the clothes we wear can be hot and humid.} \\ \hline 
PA-BART & \textcolor{blue}{In the dead of summer, in Ohio, the clothes complained and humid.} \\ \hline 
\hline
Literal Input & Streaks of electricity zipped through the air, swirling around and around dangerous (ex.10)\\ \hline
Human Ref & \textcolor{red}{Streaks of electricity cackled through the air, swirling around and around dangerous.} \\ \hline
COMET & \textcolor{orange}{Streaks of electricity stain carpet through the air, swirling around and around dangerous.} \\ \hline
BL-BART & \textcolor{violet}{Streaks of electricity were shooting through the air, swirling around and around the building.} \\ \hline 
PA-BART & \textcolor{blue}{Streaks of electricity danced around and around dangerous.} \\ \hline 
\hline
Literal Input & The sun is nowhere to be found, and the streets are eerily quiet (ex.11)\\ \hline
Human Ref & \textcolor{red}{The sun seems to be on leave, and the streets are eerily quiet.} \\ \hline
COMET & \textcolor{orange}{The sun down, and the streets full of traffic.} \\ \hline
BL-BART & \textcolor{violet}{The sun, and the streets.} \\ \hline 
PA-BART & \textcolor{blue}{The sun refuses to be found, and the streets are polite} \\ \hline 
\hline
Literal Input & The wind blew through me fast. (ex.12) \\ \hline
Human Ref & \textcolor{red}{The wind stampeded through me fast.} \\ \hline
COMET & \textcolor{orange}{The wind howl through me fast.} \\ \hline
BL-BART & \textcolor{violet}{The wind was going through me fast.} \\ \hline 
PA-BART & \textcolor{blue}{The wind ran me fast.} \\ \hline 

\end{tabular}
}
\caption{\small Additional qualitative examples for personification outputs: literal input, \textcolor{red}{human writing}, \textcolor{orange}{COMET}, \textcolor{violet}{BL-BART}, and \textcolor{blue}{PA-BART}.}
\label{appendix:qualitative_examples}
\end{table}

\end{document}